\documentclass{llncs}

\usepackage{amsmath}
\usepackage{amssymb}
\usepackage{graphicx}
\usepackage{wrapfig}
\usepackage{float}
\usepackage{subfigure}
\usepackage{capt-of}
\usepackage{cite}
\usepackage[utf8]{inputenc}
\usepackage{graphicx}
\usepackage{float}
\usepackage{wrapfig}
\usepackage{listings}
\usepackage{color}
\usepackage[autostyle]{csquotes}  

\definecolor{mygreen}{rgb}{0,0.6,0}
\definecolor{mygray}{rgb}{0.5,0.5,0.5}
\definecolor{mymauve}{rgb}{0.58,0,0.82}
\usepackage[numbers]{natbib}
\usepackage{wrapfig}
\usepackage{url}
\usepackage{algorithm}
\usepackage{algpseudocode}
\usepackage{booktabs}
\usepackage[table,xcdraw]{xcolor}

\usepackage[section]{placeins}

\newcommand{\apsis}{\textit{apsis}}

\newcommand{\argmin}{\operatornamewithlimits{argmin}}
\title{\apsis}
\subtitle{Framework for Automated Optimization of Machine Learning Hyper Parameters}

\author{Frederik Diehl \and Andreas Jauch}

\institute{Technische Universit\"at M\"unchen \\[5mm]
{\tt\small frederik.diehl@tum.de \ \ \ andreas.jauch@tum.de}}

\begin{document}
\maketitle
\section{Introduction}
Machine learning and the algorithms used for it have become more and more complex in the past years.  Especially the growth of Deep Learning architectures has resulted in a large number of hyperparameters - such as the number of hidden layers or the transfer function in a neural network - which have to be tuned to achieve the best possible performance.\\
Since the result of a hyperparameter choice can only be evaluated by completely training the model, every single hyperparameter evaluation requires potentially huge compute costs. Depending on the task, such an evaluation can take from hours to weeks, for potentially hundreds of evaluations.\\
There exist several methods to select the hyperparameters. The one most commonly used is grid search, which evaluates all combinations of chosen values for each parameter. Clearly, this scales exponentially in the parameter dimensions. Additionally, this risks evaluating parameter combinations where the only change is varying an insignificant parameter multiple times.\\
In comparison, a continuous evaluation of randomly chosen parameter values as presented in \cite{bergstraSum} avoids the latter problem, and has the advantage of being easily scalable and easy to parallelize.\\
In general, most projects use grid search or random search in combination with a human expert who determines small areas of the hyperparameter space to be automatically evaluated. This method obviously depends on expert knowledge and is therefore very difficult to scale and not transferrable.\\
Bayesian Optimization, which constructs a surrogate function using Gaussian Processes, aims to rectify this and has been shown to deliver good results on the hyperparameter optimization problem, for example in \cite{snoek2012practical}.\\

The \textbf{\apsis\ toolkit} presented in this paper provides a flexible framework for hyperparameter optimization and includes both random search and a bayesian optimizer. It is implemented in Python and its architecture features adaptability to any desired machine learning code. It can easily be used with common Python ML frameworks such as \textit{scikit-learn} \cite{scikit-learn}. Published under the MIT License other researchers are heavily encouraged to check out the code, contribute or raise any suggestions.\\
The code can be found at \url{github.com/FrederikDiehl/apsis}.

In chapter \ref{chap:bayesian} the concept of Bayesian Optimization is briefly introduced and some theoretical results \apsis\ is based on are presented. The next chapter covers \apsis ' flexible architecture followed by a chapter on performance evaluation. Finally the conclusion lists possible further steps to the take the project to the next level. 
\\
We want to thank our supervisors Prof. Dr. Daniel Cremers and Dipl. Inf. Justin Bayer for their outstanding support and helpful contributions.
\section{Bayesian Optimization for Optimizing Hyperparameters}
\label{chap:bayesian}
The general objective in hyperparameter optimization is to minimize a loss function $L$ or other performance measure of a machine learning algorithm $A$ with respect to the hyperparameter vector $\lambda$ on withheld data. 
\begin{align*}
\hat{\lambda} = \underset{\lambda \epsilon \Lambda}{\argmin} \left(\underbrace{ L(X_\text{test}, A_\lambda(X_\text{train}))}_{\Psi(\lambda)} \right)
\end{align*}
In the following the true objective function will be called $\Psi$ and the space of hyperparameters will be called $\Lambda$. \\
\begin{figure}[H]
\centering
\includegraphics[width=0.85\linewidth]{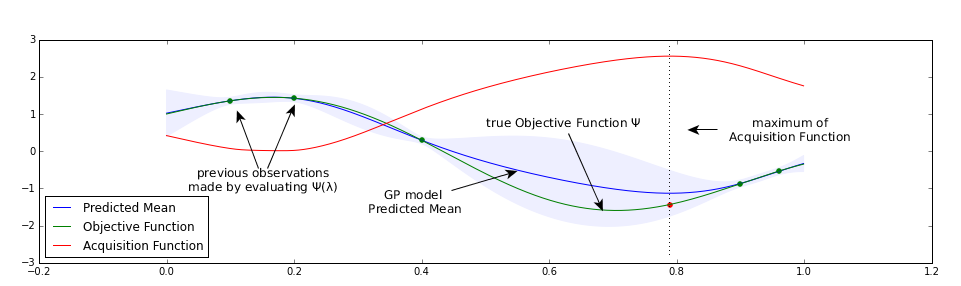}
\captionof{figure}{Concept of sequential model-based optimization using a Gaussian Process model and an Expected Improvement acquisition function, here shown for a 1D toy minimization problem.}
\label{fig:gaussOptOverview}
\end{figure}
\noindent
Since every evaluation of $\Psi$ with respect to a certain $\lambda$ incurs high cost, one goal should be to minimize their number. In Bayesian Optimization, a surrogate model is built to approximate the objective function $\Psi$. Nearly all of the Bayesian Optimization literature uses Gaussian Processes as a surrogate function. This is constructed from already evaluated samples.\\
The next candidate hyperparameter values to be evaluated is obtained from the surrogate model using a utility or acquisition function $u$.
\\
Figure \ref{fig:gaussOptOverview} illustrates these concepts. Following this procedure the problem of hyperparameter optimization now turns into the problem of maximizing the acquisition function $u$.\\

\begin{figure}
\centering
\includegraphics[width=\linewidth]{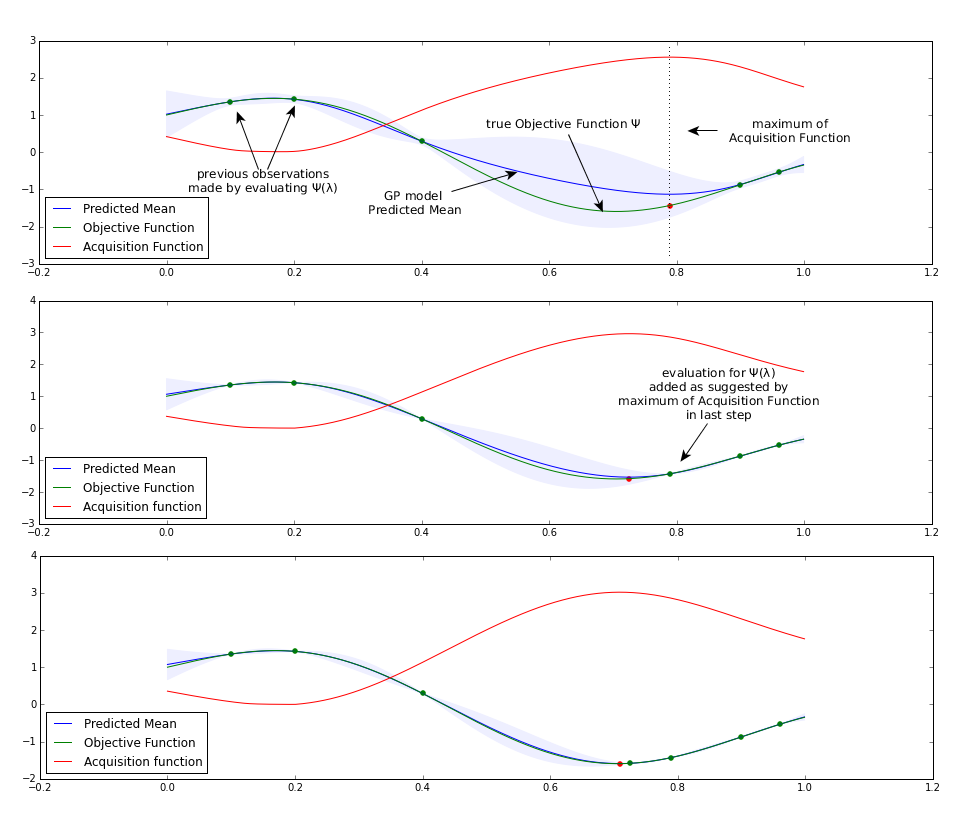}
\vspace{-1em}
\captionof{figure}{Minimization of a one dimensional function with Bayesian Optimization with a GP and Expected Improvement acquisition function. First, only a few (randomly chosen) samples are available. Now the acquisition function balances the trade-off between exploration and exploitation having its maximum to be at the point $\lambda_{t5}$. $\Psi(\lambda_{t5})$ is evaluated and the GP posterior is updated at t=6.}
\label{fig:gaussOpt}
\end{figure}

\noindent
In total, the Bayesian Optimization procedure is shown in algorithm \ref{alg:smbo}, and is also called sequential model based optimization algorithm (SMBO) in the literature. \\
In this algorithm, a model $M_t$ is used to approximate the complex response function $\Psi(\lambda)$. The history of points where $\Psi$ has been evaluated is stored in $H$. We run this algorithm for a fixed number of $T$ step. In each of the steps we first search the extremum of the acquisition function $u$ on the model $M_{t-1}$.  Afterwards, we execute the expensive evaluation of $\Psi$ at the found extremum $\lambda^{*}$ and add this evaluation to our history. Finally we use the updated history information to update $M_t$. Figure \ref{fig:gaussOpt} exemplarily shows this iterative approach for a one dimensional objective function.

\begin{algorithm}
\vspace{0.3em}
\begin{algorithmic}[1]
\Function{BayOpt}{$\Psi$, $M_0$, $T$, $u$}
\State $H \gets \emptyset$;
\vspace{0.3em}
\ForAll{$t \in \{1..T\}$}
\State $\lambda^{*} \gets argmin_\lambda \big( u(\lambda|M_{t-1})\big)$;
\State Evaluate $\Psi(\lambda^{*})$;
\State $H \gets H \cup \left(\lambda^{*}, \Psi(\lambda^{*})\right)$;
\State $M_{t} \gets $ refit surrogate model $M_{t-1}$ to updated $H$;
\EndFor
\vspace{0.3em}
\State \Return $H$;
\vspace{0.3em}
\EndFunction
\end{algorithmic}
\captionof{algorithm}{Bayesian Optimization Algorithm \cite{bergstraSum}. Notation adapted.}
\label{alg:smbo}
\end{algorithm}

\subsection{Acquisition Functions}
\label{chap:acquisitions}
One of the most important design decisions is the acquisition function used to interpret the model. Most related papers introduce the Probability of Improvement function as a starting point and historical reference, but in general, Expected Improvement works significantly better in practice. Having implemented both of these functions, \apsis\ uses Expected Improvement as a default.

\subsubsection{Probability of Improvement}
 was first shown by \citeauthor{kushner1964new} \cite{kushner1964new} and states the simple idea of maximizing the probability to achieve any improvement by choosing $\lambda$ as a next point to sample from $\Psi$.
\begin{align}
u_{\text{PI}}(\lambda | M_{t}) = \Phi \left( \frac{\mu_{M_{t}}(\lambda)-f(\lambda^{*})}{\sigma_{M_{t}}(\lambda; \theta)} \right)
\end{align}
That approach already indicates that it will favour exploitation of well-known areas over exploration of less-known regions of your optimization space $\Lambda$.

\subsubsection{Expected Improvement}
A better criterion is the expected value of the actual improvement achieved when choosing to evaluate the objective for the next $\lambda$ value.
\begin{align}
\label{eq:ei_normal}
u_{\text{EI}}(\lambda| M_{t}) = \underset{-\infty}{\int}^{\infty} \underbrace{max(y^{*} - y, 0)}_{\text{value of improvement}} \text{  }\cdot \underbrace{p_M(y|\lambda)}_\text{probability of improvement}\text{  }dy
\end{align}
Since this integral cannot be directly computed, an analytical solution is helpful. For Gaussian Processes it has been shown \cite{brochu, snoek2012practical} that the analytical solution of basic EI is obtained as
\begin{align}
\label{eq:analytic}
u_{\text{EI}}(\lambda| M_{t}) = \sigma(\lambda) \cdot \left( z(\lambda) \cdot \Phi(\lambda) + \phi(\lambda) \right)
\end{align}
using $z(\lambda)$ defined as 
\begin{align*}
z(\lambda) = \frac{f(\lambda^{*}) - \mu(\lambda)}{\sigma(\lambda)}
\end{align*}
where $\mu(\lambda)$ and $\sigma(\lambda)$ are the mean and standard deviation as given from the Gaussian Process used. $\phi(\lambda)$ and $\Phi(\lambda)$ mark the standard normal distribution density and cumulative distribution function.
\noindent
Since \apsis\ is supposed to handle both problems of maximization or minimization of the objective function both scenarios had to be incorporated into the acquisition function. Additionally, a trade-off parameter $\zeta$ to control the exploitation/exploration behaviour has been incorporated into $u$ as suggested in \cite{brochu}.
This leads to the following modified version of $z(\lambda)$ that is used in \apsis , where $MAX$ is an integer being $1$ when the objective function has to be maximized and $0$ otherwise.
\begin{align*}
z(\lambda) = \frac{(-1)^{MAX} \cdot \left( f(\lambda^{*}) - \mu(\lambda) + \zeta\right)}{\sigma(\lambda)}
\end{align*}

\subsection{Expected Improvement Optimization}
\label{chap:acquisitions_ei_opt}
One possibility for optimizing the acquisition function is to use gradient based optimization methods as they usually feature a better convergence speed than methods not relying on first order derivative information.
\subsubsection{EI Gradient Derivation}
Hence the gradient of EI had to be analytically derived.
Applying the product rule to equation (\ref{eq:analytic}) one obtains
\begin{align}
\nabla EI(\lambda) =& \text{ }\nabla \sigma(\lambda) \cdot \big(z(\lambda) \cdot \Phi(\lambda) + \phi(\lambda)\big) + \sigma(\lambda) \cdot \nabla \big(z(\lambda) \cdot \Phi(\lambda) + \phi(\lambda)\big)
 \label{eq:gradient_before}
\end{align}
\begin{multline}
\nabla EI(\lambda) = \frac{\nabla \sigma(\lambda) \cdot EI(\lambda)}{\sigma(\lambda)} \\ + \sigma(\lambda) \cdot \big( \nabla z(\lambda)\Phi(z) + \underbrace{z(\lambda)\phi(z)\cdot \nabla z(\lambda) - z(\lambda)\phi(z)\cdot \nabla z(\lambda)}_{= 0} \big)
\end{multline}
In the above $\nabla \phi(x) = -x \cdot \phi(x)$ has been used and the last two terms of the sum cancel out leading to the compact result
\begin{align}
\nabla EI(\lambda) &= \frac{\nabla \sigma(\lambda) \cdot EI(\lambda)}{\sigma(\lambda)} + \sigma(\lambda) \cdot \big( \nabla z(\lambda)\Phi(z) \big).
 \label{eq:gradient_simplified}
\end{align}
Finally, we compute the derivative of $z(\lambda)$ using the product rule.
\begin{align}
\nabla z(\lambda) &= \frac{(-1)^{MAX} \cdot - \nabla\mu(\lambda)}{\sigma(\lambda)} - \frac{(-1)^{MAX}}{{\sigma}^{2}(\lambda)} \cdot \left( f(\lambda^{*}) - \mu(\lambda) + \zeta\right) \cdot  \nabla \sigma(\lambda) \\
&= \frac{(-1)^{MAX} \cdot - \nabla\mu(\lambda)}{\sigma(\lambda)} - \frac{z(\lambda) \cdot \nabla \sigma(\lambda)}{\sigma(\lambda)}
\end{align}
It is more convenient for the implementation to have $\nabla \sigma^{2}(\lambda)$ instead of $\nabla \sigma(\lambda)$, since the GP framework used in \apsis\ returns both the mean and variance gradients. Fortunately, using the chain rule, one can derive
\begin{align}
\nabla \sigma(\lambda) = \frac{\nabla \sigma^{2}(\lambda)}{2\sigma(\lambda)} 
\end{align}
Using all of the above and inserting it in (\ref{eq:gradient_simplified}), the full EI gradient result is
\begin{align*}
\nabla EI(\lambda) &= \frac{\nabla \sigma^{2}(\lambda)}{2\sigma(\lambda)}  - (-1)^{MAX} \cdot \nabla\mu(\lambda) \cdot \Phi(z(\lambda)) -  \nabla \sigma^{2}(\lambda) \cdot \Phi(z(\lambda)) \cdot \frac{z(\lambda)}{2\sigma(\lambda)}
\end{align*}
which is the formula implemented in \apsis .
\subsubsection{Available Optimization Methods}
The following gradient and non-gradient based optimization methods are available in \apsis . 
\begin{itemize}
\item Quasi-Newton optimization using inverse BFGS \cite[p. 72]{NLO}
\item Limited Memory BFGS with bounds algorithm (default) \cite{lbfgsb}
\item Nelder-Mead method \cite{neldermead}
\item Powell method \cite{powell}
\item Conjugate Gradient method \cite{nocedal1999numerical}
\item inexact/truncated Newton method using Conjugate Gradient to approximately solve the Newton Equation \cite[p. 62]{NLO} \cite{nocedal1999numerical}
\item random search 
\end{itemize}
Except for random search the implementations of these methods are taken from the SciPy project \cite{scipy}. In order to provide these methods with a promising starting point $\lambda_0$ a random search maximization is always executed first and the best resulting $\lambda$ will be used as $\lambda_0$ for any of the further optimization methods.\\

\noindent
The BFGS method marks the strongest of these methods and offers the best convergence speed since it uses gradient information and approximates the second order information. 
Similar to the original Newton method it computes the next optimization step $s_k$ by solving the Newton equation
\begin{align}
\label{eq:Newton}
\nabla^{2}f(\lambda_k) \cdot s_k = - \nabla f(\lambda_k)
\end{align}
but as a so called Quasi Newton method it tries to approximate the Hessian $\nabla^{2}$. Therefore it maintains the Quasi Newton equation (\ref{eq:QNG}) satisfied in every step to make sure $H$ provides a well enough approximation for the Hessian.
\begin{align}
\label{eq:QNG}
H_{k+1} \cdot (\lambda^{k+1} - \lambda^{k}) = \nabla f(\lambda^{k+1}) - \nabla f(\lambda^{k})
\end{align}
Since equation (\ref{eq:Newton}) can be solved for $s_k$ by inverting the Hessian, the inverse BFGS method used in \apsis\  directly approximates the inverse of the Hessian using an iterative algorithm to speed up optimization. \\
It will however not always converge since the algorithm involves a division by something that can become $0$ \cite[p. 72]{NLO}. In that case the algorithm stops gracefully with an error and \apsis\ relies on the random search result. Since the optimization space is bounded in real settings \apsis\ by default uses an adopted version of BFGS, the L-BFGS-B \cite{lbfgsb} algorithm. In contrast to ordinary BFGS it does not store the full $n\times n$ Hessian but only stores a few vectors to implicitly represent the Hessian approximation $H$. Furthermore it has been extended by \citeauthor{lbfgsb} \cite{lbfgsb} to respect simple bounds constraints for each variable definition which is sufficient for the bounds used in \apsis .
\section{Architecture}
\subsection{General Architectural Overview}
\begin{figure}
\centering
\includegraphics[width=\linewidth]{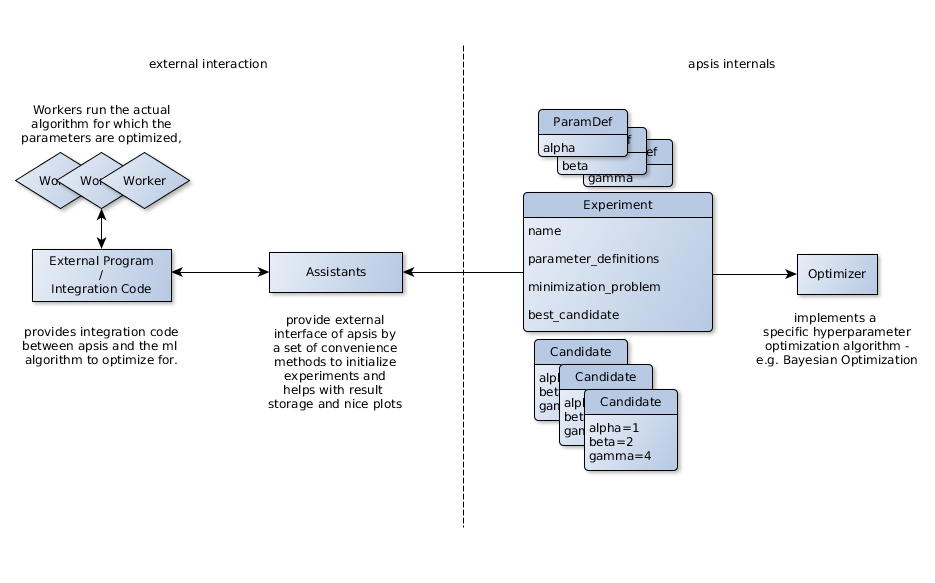}
\captionof{figure}{\apsis\ general architectural overview}
\label{fig:interoperability}
\end{figure}

The architecture of \apsis\ is designed to be interoperable with any Python machine learning framework or self-implemented algorithm. \apsis\ features an abstraction layer for the underlying optimization framework as depicted in figure \ref{fig:interoperability}. Every optimizer adheres to the abstract base class \texttt{Optimizer}. The optimizer uses three important model classes to control optimization. \texttt{Candidate} is used to represent a specific hyperparameter vector $\lambda$. \texttt{Experiment}s are used to define a hyperparameter optimization experiment. An \texttt{Experiment} holds a list of \texttt{ParamDef} objects that define the nature of each specific hyperparameter. The external program interacts with \apsis\ through a set of \textit{Assistants}. They help to initialize {\apsis'} internal structure and models and provide convenient access to optimization results, and can optionally store and plot them or compare several experiments. Additionally, the external program has to provide some integration code between the external program's machine learning algorithm and the \apsis\ interface. Since the external program could be multi-threaded or clustered \apsis\ refers to the processes running the machine learning algorithms as \textit{Workers}.

\noindent
On the implementation level \apsis\ is made up by six packages reflecting this architecture. Table \ref{tab:package_structure} lists all of them and briefly describes their purpose.


\begin{table}[h]
\resizebox{\textwidth}{!}{%
\begin{tabular}{|l|l|l|}
\hline
\rowcolor[HTML]{000000} 
{\color[HTML]{FFFFFF} \textbf{Package}} & {\color[HTML]{FFFFFF} \textbf{Description}}                                                                                                                   & {\color[HTML]{FFFFFF} \textbf{Contents}}                                                                                                                                          \\ \hline
assistants                              & \begin{tabular}[c]{@{}l@{}}Contains lab and experiment assistants for \\ \apsis' external interface.\end{tabular}                                              & \begin{tabular}[c]{@{}l@{}}experiment\_assistant.py\\ lab\_assistant.py\end{tabular}                                                                                              \\ \hline
optimizers                              & \begin{tabular}[c]{@{}l@{}}Contains all available optimizers and the abstract \\ base class. Contains a subpackage for \\ Bayesian Optimization.\end{tabular} & \begin{tabular}[c]{@{}l@{}}optimizers.py\\ random\_search.py\\ bayesian\_optimization.py\\ bayesian/acquisition\_functions.py\end{tabular}                                        \\ \hline
models                                  & Contains the model classes.                                                                                                                                   & \begin{tabular}[c]{@{}l@{}}candidate.py\\ experiment.py\\ parameter\_definition.py\end{tabular}                                                                                   \\ \hline
tests                                   & Contains unit tests for complete \apsis\ code.                                                                                                                  & \begin{tabular}[c]{@{}l@{}}test\_assistants/...\\ test\_models/...\\ test\_optimizers/...\\ test\_utilities/...\end{tabular}                                                      \\ \hline
utilities                               & Utility functions used accross all packages.                                                                                                              & \begin{tabular}[c]{@{}l@{}}benchmark\_functions.py\\ file\_utils.py\\ import\_utils.py\\ logging\_utils.py\\ optimizer\_utils.py\\ plot\_utils.py\\ randomization.py\end{tabular} \\ \hline
demos                                   & \begin{tabular}[c]{@{}l@{}}Some demos intended for learning how to \\ use \apsis\ by examples.\end{tabular}                                                     & \begin{tabular}[c]{@{}l@{}}demo\_MNIST.py\\ demo\_MNIST\_MCMC\_py\end{tabular}                                                                                                    \\ \hline
\end{tabular}
}
\vspace{0.5em}
\caption{list of \apsis\ packages with their purpose and contents}
\label{tab:package_structure}
\end{table}

\subsection{Model Objects}
\subsubsection{Candidate} represents a specific hyperparameter vector $\lambda$ and optionally stores the result of the objective function $\Psi(\lambda)$ achieved under this $\lambda$ if already evaluated. Additionally it can store the cost occurred for evaluation and any other meta information used by the worker, e.g. the worker could keep track of the classifier's weights in each step and store them there.
The actual vector $\lambda$ is stored as a dictionary such that every parameter dimension $\lambda^{(i)}$ is named.

\subsubsection{Experiment} stores information about the nature of the parameters to be optimized, defines if the problem is for minimization and maximization. It keeps track of successfully evaluated, currently evaluated and to be evaluated \texttt{Candidate}s. It stores the best \texttt{Candidate} and optionally names the experiment.\\
Additionally, it provides methods for semantic check if candidates are valid for this experiment, and can convert itself to a CSV format for result storage.

\subsubsection{ParamDef} 
\begin{figure}[H]
\centering
\includegraphics[width=0.9\linewidth]{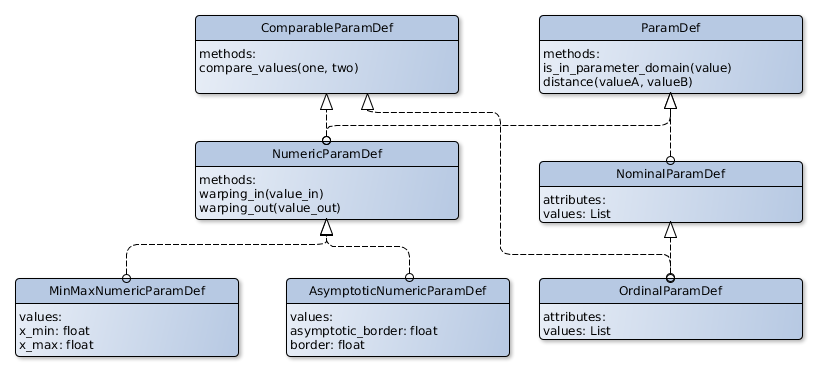}
\captionof{figure}{Overview of ParamDef model classes and their relations.}
\label{fig:param_defs}
\end{figure}

is the most general superclass used to define the nature of one hyperparameter dimension. It makes no assumption on the nature of the stored parameter. \texttt{NominalParamDef} defines non-comparable, unordered parameters and stores all available values in a list. \texttt{OrdinalParamDef} represents ordinal parameters where the order is maintained by the position in the list of values. As ordinal parameters can be compared they adhere to the \texttt{ComparableParamDef} interface and provide a \texttt{compare\_values} function that provides similar semantic to the Python integrated \texttt{\_\_cmp\_\_} function.
Furthermore, \apsis\ comes with special support for numeric parameters represented by the \texttt{NumericParamDef} class. Numeric param defs are comparable and are represented internally by a warping into the $[0,1]$ domain. Hence they need to be given an inwards and outwards warping function. The warped parameters are now compared and treated according to the rules of treating ordinary floats between $0$ and $1$. To ease initialization of numeric parameters \texttt{MinMaxNumericParamDef} automatically provides a warping assuming an equal distribution of the parameter between the given minimal and maximal value. \texttt{AsymptoticNumericParamDef} provides a parameter definition for a parameter which is expected to be close to one value. For example, learning rates can be estimated as being close to 0. This leads to significantly better results during optimization, as long as some expert knowledge is available.
\\Figure \ref{fig:param_defs} depicts the structure of available parameter types and there inheritance relationships.

\subsection{Optimization Cores}
The abstract base class \texttt{Optimizer} defines the common interface of any optimization algorithm provided in \apsis . Central is the method \texttt{get\_next\_candidates}. Based upon information stored in the given experiment instance the method provides its user with one or several promising candidates to evaluate next. On construction the optimizer can receive a dictionary of optimizer specific parameters to define optimization related hyperparameters. Note that these parameters are not the ones that are subject of optimization but are parameters to govern the optimization behaviour such as which acquisition function or kernel is used in Bayesian Optimization.
\apsis\ provides two different optimization cores: a simple random search based optimizer called \texttt{RandomSearch} and the \texttt{SimpleBayesianOptimizer}.

\subsubsection{RandomSearch}
Implements a very simple random search optimizer. For parameters of type \texttt{NumericParamDef} a uniform random varibale between $0$ and $1$ is generated to select a value in warped space for each parameter. For parameters of type \texttt{NominalParamDef} a value is drawn uniformly at random from the list of allowed values. All random numbers are generated using the numpy \cite{numpy} random  package.

\subsubsection{SimpleBayesianOptimizer}
The \texttt{SimpleBayesianOptimizer} works according to the theory described in chapter \ref{chap:bayesian}. It is called simple since it currently works with one concurrent worker at a time only (though  a multi worker variant is planned). It uses Gaussian Processes and their kernels provided by the GPy framework \cite{gpy2014}. For the acquisition functions implemented and the methods in use for their optimization it shall be referred to chapter \ref{chap:acquisitions} and \ref{chap:acquisitions_ei_opt}.

\subsection{Experiment Assistants}
The \texttt{BasicExperimentAssistant} provides the simplest interface between \apsis\ and the outside world. It administers at most one experiment at a time. An experiment needs to be initialized by at least specifying a name to identify it, the \texttt{Optimizer} to be used and a list containing one \texttt{ParamDef} object for each hyperparameter to be optimized. It holds and manages the \texttt{Optimizer} and \texttt{Experiment} instances and provides an abstraction layer to their interface.\\
\begin{figure}[H]
\centering
\includegraphics[width=\linewidth]{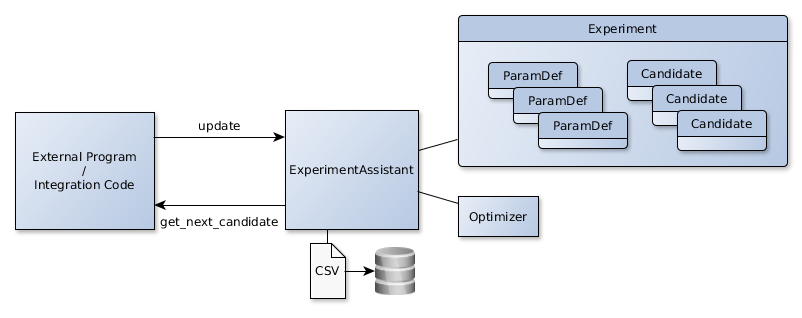}
\captionof{figure}{The role of experiment assistants in \apsis .}
\label{fig:exp_assistant_architecture}
\end{figure}
\noindent
The experiment assistants provide the two methods \texttt{get\_next\_candidate} and \texttt{update} that should be called by the external program when it is ready to evaluate a new candidate or wants to notify \apsis\ about work on a \texttt{Candidate}. The \texttt{Candidate} doesn't necessarily need to be one which was provided by \apsis\ but can be any Candidate which adheres to the parameter space.
Furthermore \texttt{BasicExperimentAssistant} cares for storing the results to CSV files when running to be sure to have all information available after termination or abortion of an experiment run. The behaviour of the CSV writing can be controlled upon initialization of the experiment assistant. 
As an extension \apsis\ provides \texttt{PrettyExperimentAssistant} that can additionally create nice plots on the experiment. Chapter \ref{chap:evaluation} shows some of these plots.

\subsection{Lab Assistants}
\begin{figure}[H]
\centering
\includegraphics[width=\linewidth]{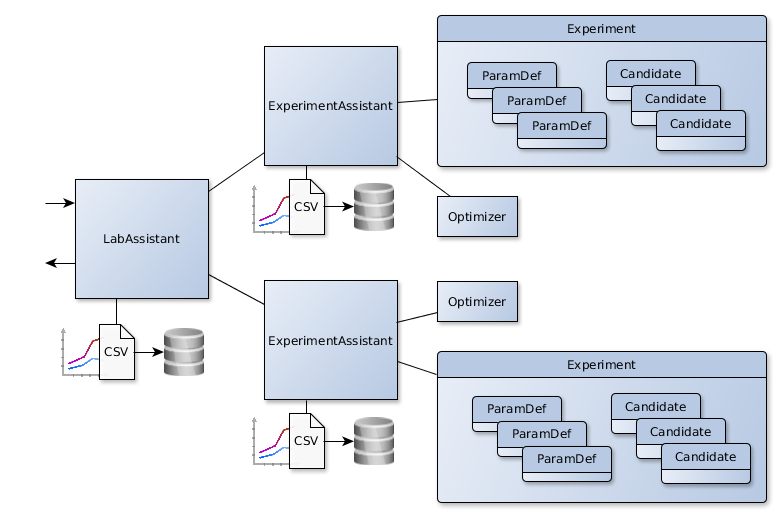}
\captionof{figure}{Using lab assistants to manage and compare several experiments at once.}
\label{fig:lab_assistant_architecture}
\end{figure}
\texttt{BasicLabAssistant} deals with the requirement to run and control several experiments at once. It holds a dictionary of experiment assistants and provides a common interface to all of them following the same semantic as the experiment assistants themselves. The advantages of using lab assistants in contrast to controlling several experiments at the same time manually is that the results can now be stored together and compared. \texttt{BasicLabAssistant} provides only that function, while \texttt{PrettyLabAssistant} adds comparative plots for the comparison of multiple optimizers.

\section{Evaluation}
\label{chap:evaluation}
This section evaluates \apsis\ on several benchmark and one real world example. All experiments are evaluated using cross validation and $10$ initial random samples that are shared between all optimizers in an experiment to ensure comparability.

\subsection{Evaluation on Branin Hoo Benchmark Function}
Some recent papers on Bayesian Optimization for machine learning publish evaluations on the Branin Hoo optimization function \cite{snoek2012practical}. The Branin Hoo function 

\begin{align*}
f_\text{Branin}(x) = a \cdot (y - b\cdot x^{2} + c \cdot x - r)^{2} + s \cdot (1-t) \cdot cos(x) + s
\end{align*}
using values proposed in \cite{simulationlib} is defined as
\begin{align*}
f_\text{Branin}(x) = (y - \frac{5.1}{4 \pi^{2}}\cdot x^{2} + \frac{5}{\pi} \cdot x - 6)^{2} + 10 \cdot (1- \frac{1}{8 \pi}) \cdot cos(x) + 10.
\end{align*}
In contrast to our expectations Bayesian Optimization was not able to outperform random search on Branin Hoo. Still the result is much more stable and the bayesian optimizer samples only close to the optimum.

\begin{figure}[H]
\centering
\begin{minipage}[b]{0.39\linewidth}
\centering
\includegraphics[width=0.97\linewidth]{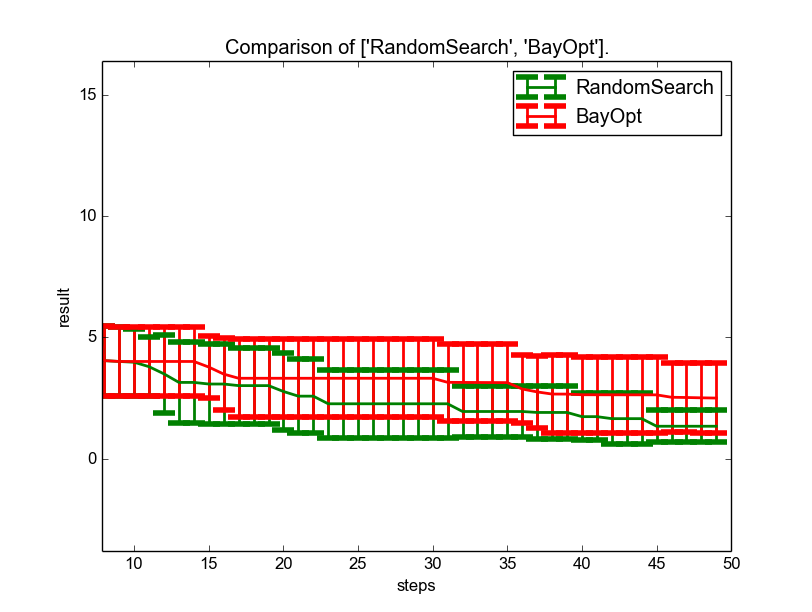}
\label{fig:minipage1}
\end{minipage}
\begin{minipage}[b]{0.6\linewidth}
\centering
\includegraphics[width=0.97\linewidth]{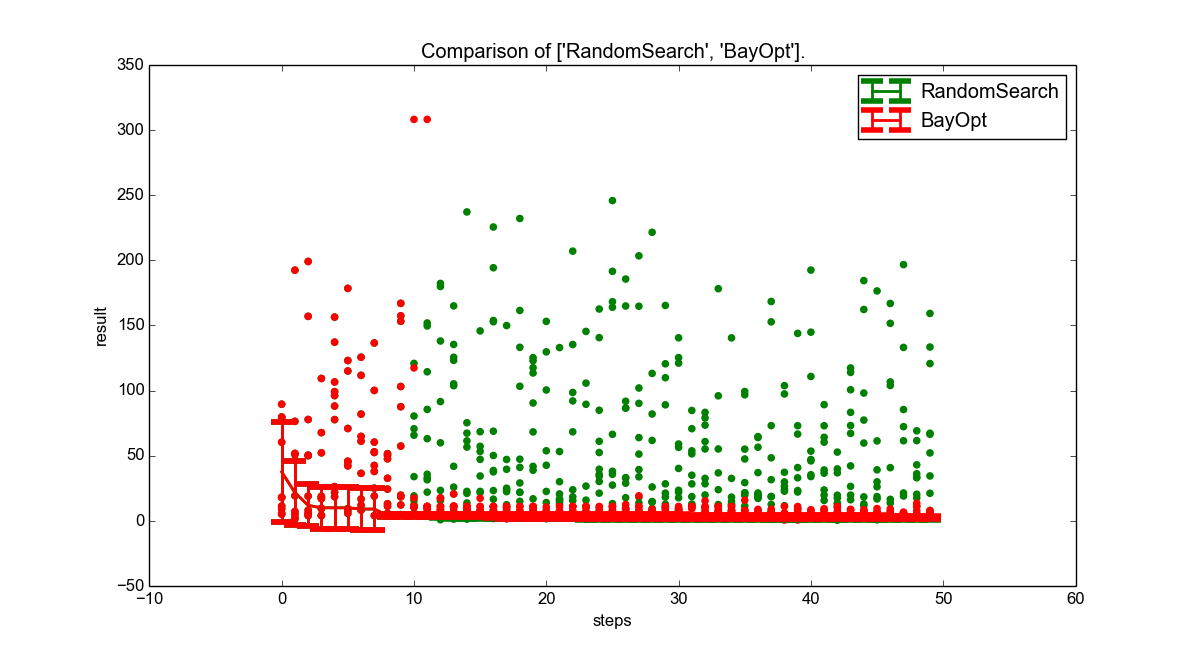}
\end{minipage}
\captionof{figure}{Comparison of Bayesian Optimization vs. random search optimization on Branin Hoo function. The left figure shows the best result in every step. Here, random search clearly outperforms Bayesian Optimization. The right plot additionally plots each function evaluation as a dot. Here, it is apparent that Bayesian Optimization works a lot more stable and does not evaluate as many non-promising candidates as random search.}
\label{fig:branin_eval}
\end{figure}

\subsection{Evaluation on Multidimensional Artificial Noise Function}

\begin{figure}
\begin{minipage}[b]{1\linewidth}
\centering
\includegraphics[width=0.97\linewidth]{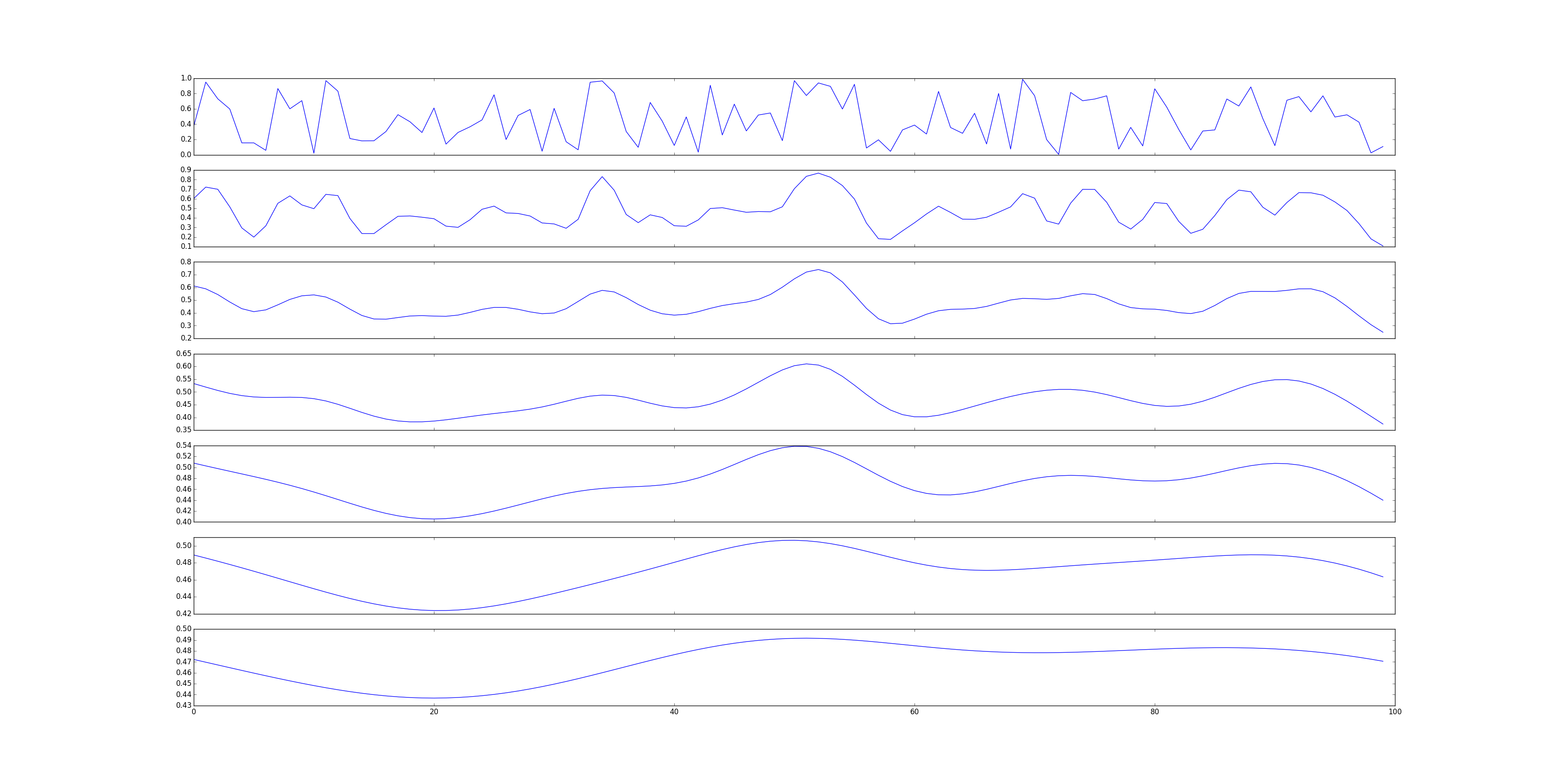}
\captionof{figure}{Plot of artificial noise function used as an optimization benchmark in \apsis. This is generated using a grid of random values smoothed over by a gaussian of varying variance.}
\label{fig:noise_function}
\end{minipage}
\centering
\begin{minipage}[b]{1\linewidth}
\centering
\includegraphics[width=0.97\linewidth]
{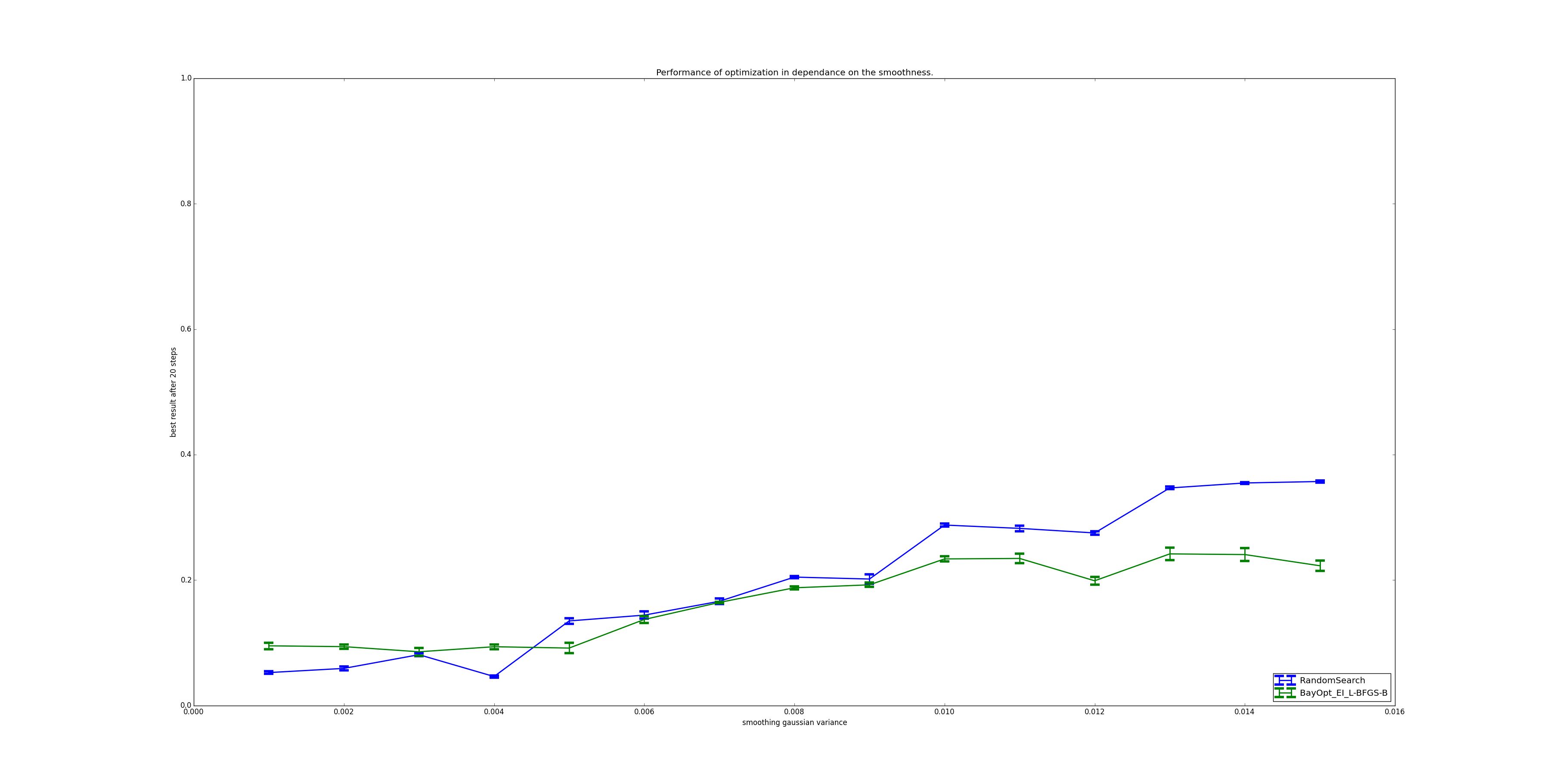}
\captionof{figure}{Plot of the end result after 20 optimization steps on a 3D artificial noise problem depending on the smoothing used. Values to the right are for smoother functions. A lower result is better.}
\label{fig:noise_perf}
\label{fig:minipage2}
\end{minipage}
\end{figure}

Compared to random search an intelligent optimizer should be better on less noisy function than on very noisy functions in theory. A very noisy function has a tremendous amount of local extrema making it hard to impossible for Bayesian Optimization methods to outperform random search. To investigate this proposition an artificial multidimensional noise function has been implemented in \apsis\ as shown in figure \ref{fig:noise_function}.\\
Using this noise function, one can generate multi-dimensional noises with varying smoothness. The construction process first constructs an $n$-dimensional grid of random points, which remains constant under varying smoothness. Evaluating a point is done by averaging the randomly generated points, weighted by a gaussian with zero mean and varying variance\footnote{Actually, only the closest few points are evaluated to increase performance.}. This variance influences the final smoothness. A one-dimensional example of generated functions for differing variances is seen in figure \ref{fig:noise_function}.\\
The result can be seen in figure \ref{fig:noise_perf}. As expected, Bayesian Optimization outperforms random search for smoother functions, while achieving a rough parity on rough functions.

\subsection{Evaluation on Neural Network on MNIST}

\begin{figure}
\centering
\includegraphics[width=1\linewidth]{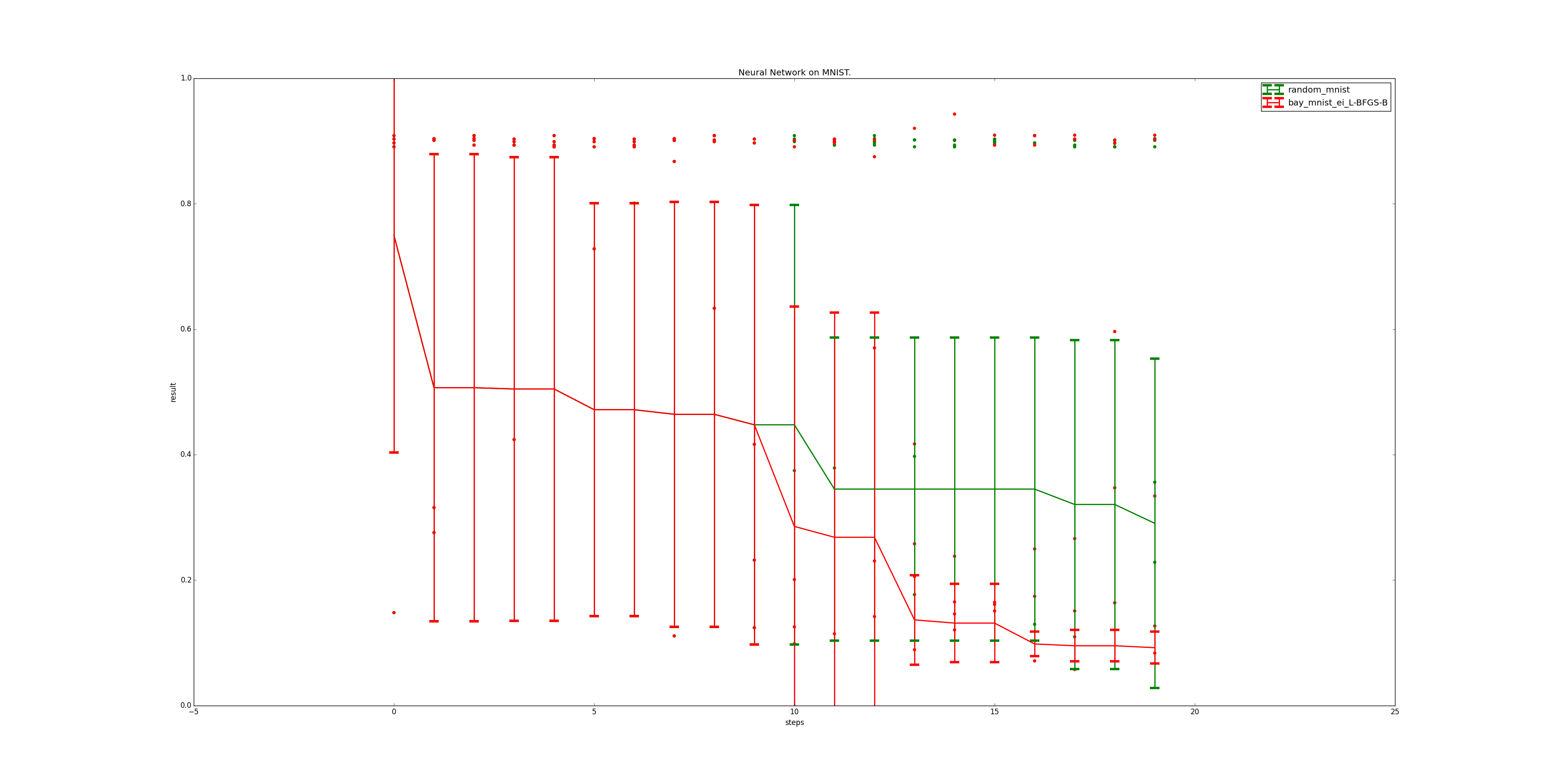}
\vspace{-1em}
\captionof{figure}{Comparison of random search and Bayesian Optimization in the context of a neural network. Each point represents one parameter evaluation of the respective algorithm. The line represents the mean result of the algorithm at the corresponding step including the boundaries of the 75\% confidence interval.}
\label{fig:nn_uniform}
\end{figure}

To evaluate the hyperparameter optimization on a real world problem, we used it to optimize a neural network on the MNIST dataset \cite{mnistlecun}. We used Breze\cite{breze2015} as a neural network library\footnote{Breze uses theano \cite{bergstra+al:2010-scipy, Bastien-Theano-2012} in the background.} in Python.\\
The network is a simple feed-forward neural network with 784 input neurons, 800 hidden neurons and 10 output neurons. It uses sigmoid units in the hidden layers, and softmax as output. We learn over 100 epochs. These parameters stay fixed throughout the optimization.\\
For assigning the neural network weights, we use a backpropagation algorithm. 
Its parameters - step\_rate, momentum and decay - are optimized over, as is $c_{wd}$, a weight penalty term. Hence, resulting in a four dimensional hyperparameter optimization.\\
We ran all neural network experiments with a five-fold cross validation. Even so, total evaluation time ran close to 24 hours on an Nvidia Quadro K2100M.\\
Figure \ref{fig:nn_uniform} shows the performance of the optimizers for each step. As can be seen, Bayesian Optimization - after the first ten, shared steps, rapidly improves the performance of the neural network by a huge amount. This is significantly more stable than the random search optimizer it is compared with.\\

\begin{figure}
\centering
\includegraphics[width=1\linewidth]{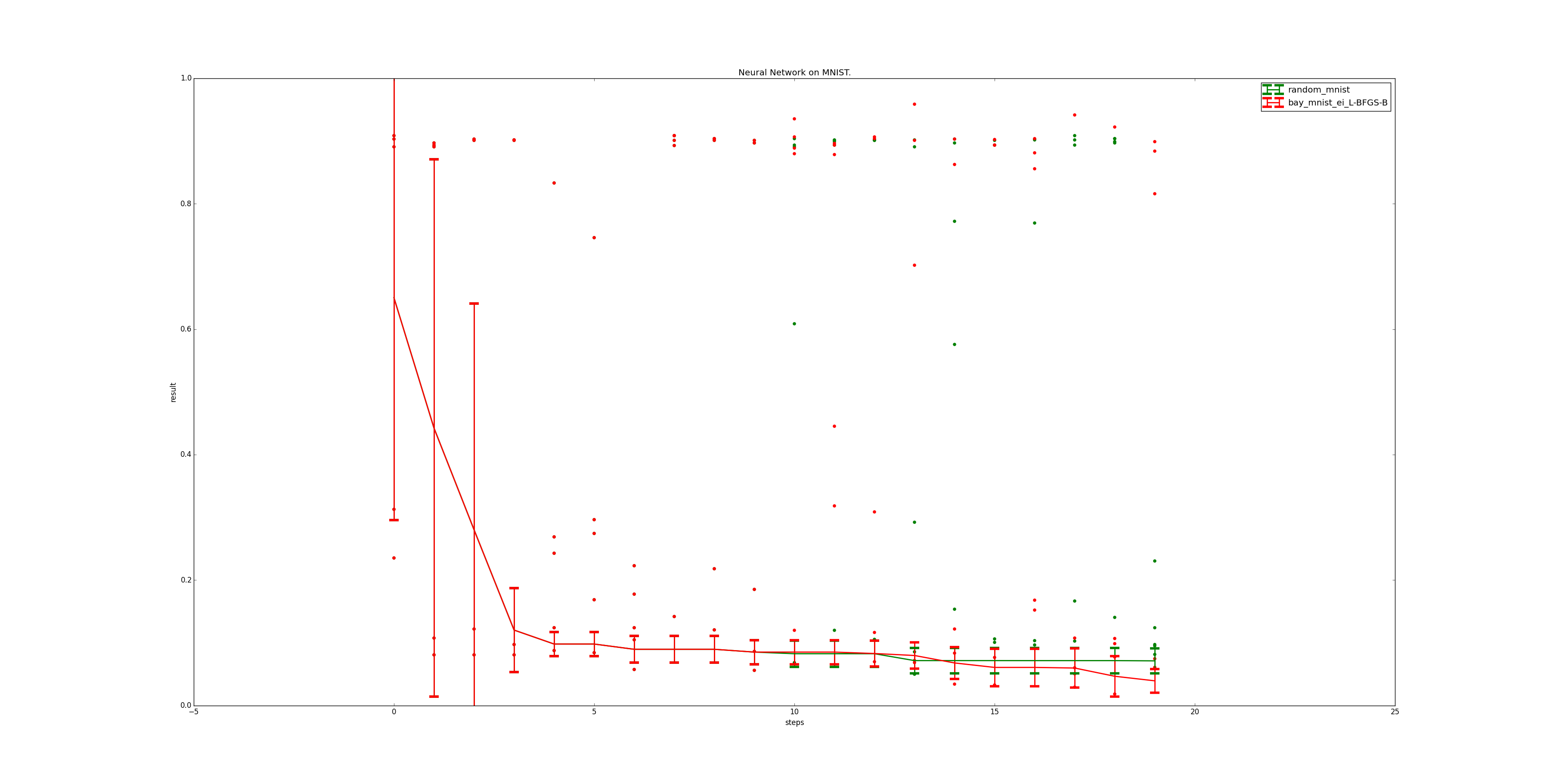}
\vspace{-1em}
\captionof{figure}{Comparison of random search and Bayesian Optimization in the context of a neural network. This optimization uses additional knowledge in that step\_rate is assumed to be close to 0 and decay to be close to 1.}
\label{fig:nn_asymp}
\end{figure}

However, the optimization above uses no previous knowledge of the problem. In an attempt to investigate the influence of such previous knowledge, we then set the parameter definition for the step\_rate to assume it to be close to 0, and the decay to be close to 1. This is assumed to be knowledge easily obtainable from any neural network tutorial.\\
The effects of this can be seen in figure  \ref{fig:nn_asymp}, and are dramatic. First of all, even random search performs significantly better than before, reaching a similar value as the uninformed Bayesian Optimization. Bayesian optimization profits, too, and decreases the mean error by about half.\\

\section{Conclusion}
With \apsis\ we presented a flexible open source framework for optimization of hyperparameters in machine learning algorithms. It implements most of the state of the art on hyperparameter optimization of current research and is open for further extension.\\
It is our hope that \apsis\ will continue to be expanded, and will be used extensively in academia.

The performance evaluation justifies that Bayesian Optimization is significantly better than random search on a real world machine learning problem. Furthermore the need for an efficient parameter optimization arises already on an experiment with only fifteen minutes of computation time per evaluation. These settings will certainly be met by any practical machine learning problem. Including very simple prior knowledge of the algorithm leads to another significant performance improvement.

There are several areas in which \apsis\ can be further improved. We would like to implement support for multiple concurrent workers, allowing us to optimize on clusters. Adding a REST web-interface allows an easy integration of \apsis\ to arbitrary languages and environments.\\
There also remain possible improvements on the implemented bayesian optimizer itself. \cite{swersky2014freeze} propose using early validation to abort the evaluation of unpromising hyperparameters. \cite{snoek2014input} propose learning input warpings, which would further reduce the reliance on previous knowledge.  \cite{snoek2012practical} use a cost function to minimize total computing time instead of the number of evaluations. The problem generally known as a tree-structured configuration space as pointed out by \cite{bergstraSum} could be tackled, allowing us to mark parameters as unused for a certain evaluation. Student-t processes might be investigated as a replacement for gaussian processes. Lastly, adding efficient support for nominal parameters in Bayesian Optimization is important for a good performance, though since no publication on that topic exists so far this might be one of the biggest barriers to tackle.

\newpage

\bibliography{literatur}{}
\bibliographystyle{plainnat}


\end{document}